\documentclass{article}

\usepackage{graphicx}
\usepackage{amsmath}
\usepackage{amssymb}
\usepackage{subcaption}
\usepackage{wrapfig}
\usepackage{multirow}
\usepackage[final]{corl_2019} % Uncomment for the camera-ready ``final'' version

\title{Hybrid system identification\\ using switching density networks}

\author{
  Michael Burke\\
  School of Informatics\\
  University of Edinburgh\\
  \texttt{michael.burke@ed.ac.uk} \\
  \And
  Yordan Hristov \\
  School of Informatics \\
  University of Edinburgh \\
  \texttt{yordan.hristov@ed.ac.uk} \\
  \And
  Subramanian Ramamoorthy \\
  School of Informatics \\
  University of Edinburgh \\
  \texttt{s.ramamoorthy@ed.ac.uk} \\
}

\begin{document}
\maketitle

%===============================================================================

\begin{abstract}

Behaviour cloning is a commonly used strategy for imitation learning and can be extremely effective in constrained domains. However, in cases where the dynamics of an environment may be state dependent and varying, behaviour cloning places a burden on model capacity and the number of demonstrations required. This paper introduces switching density networks, which rely on a categorical reparametrisation for hybrid system identification. This results in a network comprising a classification layer that is followed by a regression layer. We use switching density networks to predict the parameters of hybrid control laws, which are toggled by a switching layer to produce different controller outputs, when conditioned on an input state. This work shows how switching density networks can be used for hybrid system identification in a variety of tasks, successfully identifying the key joint angle goals that make up manipulation tasks, while simultaneously learning image-based goal classifiers and regression networks that predict joint angles from images. We also show that they can cluster the phase space of an inverted pendulum, identifying the balance, spin and pump controllers required to solve this task. Switching density networks can be difficult to train, but we introduce a cross entropy regularisation loss that stabilises training.
%Mixture density networks have become a widely used technique for regression when output data is multi-modal. Unfortunately these networks lack interpretability, are vulnerable to mode collapse, and tend to construct output distributions in an un-intuitive fashion. This paper introduces switching density networks, which rely on categorical reparametrisation for switching state space model learning. This results in a network comprising a classification layer that is followed by a regression layer. Switching density networks are constructed using a neural network that predicts the parameters of a single distribution, which is toggled by a switching layer to produce different outputs. This work shows how switching density networks can be used for hybrid system identification in a variety of tasks. We show how switching density models can be used for goal identification in a manipulation task, successfully identifying the key joint angle goals that make up an inspection task, while simultaneously learning image-based goal classifiers and regression networks that predict joint angles from images. We also show that they can identify the phase space of an inverted pendulum, and produce more robust control than a fully connected neural network when used for behaviour cloning.  Switching density networks can be difficult to train, but we show empirically that batch normalisation and a cross entropy regularisation loss stabilises training.
\end{abstract}

\keywords{Behaviour cloning, switching density networks, hybrid systems} 

%===============================================================================

\section{Introduction}

Behaviour cloning is a commonly used technique in learning from demonstration or imitation learning. Here, demonstrations of successful behaviours are used to train models that replicate the demonstrated behaviour \citep{Pomerleau,Bagnell-2015-5921}. Supervised learning problems such as these are often formulated as classification or regression tasks. The former typically assumes that some prior categorisation has been done, possibly through an initial clustering phase, while the latter tends to ignore these aspects and focuses only on predicting some real-valued output. However, many learning problems require that a hierarchical model be learned, where some latent symbolic aspect of a problem maps to a real valued observation. 

This is particularly true in robotics applications, where hierarchical learning is often key to robust, generalisable control, and learned skills are typically required to be decomposable for re-use in other applications. Historically, this problem has been addressed in a multi-stage process. For example, in the context of learning from demonstration, behaviours or low-level skills are often first identified through a clustering process, before being grounded through some learning process \citep{levine2014learning,niekum2015learning}. Hybrid systems \cite{goebel2009hybrid} offer a rich mechanism for expressing behaviours, but are typically obtained by hand. More recently, differentiable parametrisations have been exploited for gradient-based inference in hierarchical latent variable models \citep{pmlr-v32-kingma14,jang2016categorical}. This has paved the way to incorporating valuable structure into end-to-end learning models. The incorporation of structure into neural models, while still subject to debate, has gained in popularity recently \citep{penkov2019learning,karkus2019differentiable,pmlr-v97-kipf19a}, motivated as a mechanism for interpretable learning and as a means of improving performance. Interpretable robot behaviour is key if learning robots are to be trusted in practical settings.

This work introduces switching density networks (SDN) for switching control law identification. Switching density networks are a form of mixture density network \cite{bishop1994mixture} leveraging Gumbel-Softmax \citep{jang2016categorical} gating to learn to generate densities based on a binary encoded bottleneck layer. This forces the network to perform an intermediate clustering phase prior to making output predictions, which results in more interpretable, hierarchical models that allow for further reasoning. Unlike traditional hybrid system identification, which typically occurs in simple state spaces \cite{paoletti2007identification}, switching density networks allow for hybrid system identification with richer sensor data such as images.

This interpretability is a particularly useful property in robotics, and can be used to learn compositional control strategies from demonstration. In our experiments, SDNs are used to predict the controller parameters and reference states for a family of proportional-integral-derivative (PID) control laws. Hybrid systems of these control laws are applied ubiquitously across domains \cite{lunze2009handbook} including process control and automotive applications, so this family covers a broad class of applications. Experimental results show that switching density networks are able to identify the robot joint angle goals that constitute a demonstration sequence and can successfully learning visual grounding of these goals. Moreover, we show that SDNs learning PID control law families can identify the state-space regions required for pumping, spinning and balancing an inverted pendulum.  

Training such SDNs can be challenging, and we show that they are indeed vulnerable to mode collapse. This work introduces a cross-entropy batch regularisation loss that remedies this and allows for reliable training.

Formally, our goal is to identify hybrid systems or state space models conditioned on a discrete switching process, that is, to find a mapping from observation $\mathbf{z}_t$ to a state of interest, $\mathbf{x}_t$, in a stochastic dynamical system conditioned on a discrete latent variable $i_t$,
\begin{align}
q(\mathbf{x}_t|\mathbf{z}_t) &= p(\mathbf{x}_t|\mathbf{z}_t, i_t)\\
i_t &\sim p(i_t|i_{t-1}).
\end{align}

 State space models of this form are particularly powerful, as they can be used to express complex motions and behaviours using interpretable sub-components In the context of sensorimotor control for robotics, let $\mathbf{z}_t$ denote sensor data captured at time $t$ and $\mathbf{x}_t$ the pose or configuration of a robot at time $t$. $i_t$ is a discrete indicator variable controlling the transition between robot behaviour states or environment dynamics.

\section{Related work}

Numerous models and approaches \cite{MurJoh97} have been developed to address the learning problem formulated above. Gaussian mixture models fit using expectation maximisation \cite{dempster1977maximum} are widely used for clustering, while their switching state space analog, Gaussian emission hidden Markov models have a long history of application in sequence learning. Although typically fit using the Baum-Welch algorithm \cite{rabiner1986introduction} (a form of expectation maximisation), variational approaches have also been proposed for a broader class of switching state space models \cite{ghahramani2000variational}.

Learning for switching state space models can also be considered from a changepoint detection perspective, and a range of numerical inference techniques have been used to detect changepoints in sequential data \cite{ruanaidh1994recursive}. More recently, variational and gradient-based inference strategies for Bayesian learning have proved useful in hierarchical modelling \cite{pmlr-v32-kingma14,jang2016categorical} and variational auto-encoding \cite{kingma2013auto}. 

Hierarchical modelling is an effective means of incorporating structure into a learning problem, so as to avoid sample inefficient learning and improve generalisation through abstraction. Work on options learning \cite{sutton1999between, konidaris2009skill} and skill identification \cite{Niekum11,ranchod2015nonparametric} has paid significant attention to hierarchical learning, but has been a particular challenge for visuomotor control. 

Our work is inspired by sequential composition theories in robotics \cite{Burridge99}, where tasks are solved by moving between sub-controllers lying within the domains of one another. Here, we seek to identify the sub-controllers required for a given task in an end-to-end fashion, from demonstration sequences. Learning from demonstration (LfD) \cite{ARGALL2009469} is widely acknowledged as a particularly useful paradigm for robot programming. Significant progress has been made in LfD, moving beyond the direct replication of motions to produce more robust approaches \cite{atkeson1997robot} through the introduction of more general schemes for modelling motion like dynamic motion primitives \cite{pastor2009learning}, linear dynamical attractor systems \cite{Dixon04}, sparse online Gaussian processes \cite{Butterfield10,grollman2008sparse} or conditionally linear Gaussian models \cite{Chiappa10,levine2014learning} that can be used for trajectory optimisation. It is important to note that each of these systems behaves as a hybrid system, decomposing a state space into specific regions, and learning appropriate dynamics for each region. 

More recently, trajectory optimisation approaches have been extended to incorporate end-to-end learning, demonstrating robust task level visuomotor control \cite{levine2016end} through guided policy search, or using deep dynamic motion primitives \cite{8246874}. End-to-end learning has allowed for the use of domain transfer to facilitate one-shot learning \cite{Yu-RSS-18} from human video demonstrations, and for the use of reinforcement learning to learn optimised control policies \cite{Rajeswaran-RSS-18,Zhu-RSS-18}. Unfortunately, end-to-end learning approaches typically lack interpretability and are difficult to verify without policy distillation \cite{bastani2018verifiable}. \citet{burke2019explanation} fit a sequence of proportional control laws to end-to-end model demonstrations using particle filters, in an attempt to obtain a more interpretable control system, but this approach is vulnerable to performance loss if important properties of the network fail to be inferred. In contrast, this paper shows that it is possible to learn switching proportional control laws in an end-to-end fashion, by emdedding this structure into the learning process.

In computer vision, spatial transformers \cite{Jaderberg} and capsule networks \cite{sabour2017dynamic} embed learnable structured transformations in an attempt to better capture the relational properties of image attributes in convolutional neural networks. Without this structure, convolution neural networks can learn jumbled image representations \cite{hinton2011transforming}. This work shows that mixture density networks suffer from a similar problem, which switching density networks address. Switching density networks are conceptually similar to the stochastic neural network architecture proposed by \citet{Florensa17}, which uses a switching structure to learn reusible skills in a reinforcement learning setting. Our work differs by considering the use of switching structures for parameter prediction for state space models, thereby incorporating known controller structure into the learning process in a lightly supervised manner.

Switching density networks are closely related to mixture density networks \cite{bishop1994mixture}, a family of neural network constructed using $K$ output distributions. In the Gaussian mixture case, MDNs fit a weighted combination of Gaussian distributions,
\begin{align}
q(\mathbf{x}_t|\mathbf{z}_t) &= \sum_{i=1}^K \pi_i(\mathbf{z}_t) \mathcal{N}(\mathbf{x}_t|\mathbf{\mu}(\mathbf{z}_t),\mathbf{\Sigma}(\mathbf{z}_t)),
\end{align}
 using mean $\mathbf{\mu}(\mathbf{z}_t)$, variance $\mathbf{\Sigma}(\mathbf{z}_t)$ and normalised weight parameters $ \pi_i(\mathbf{z}_t)$, which are predicted using a neural network. Unfortunately, there is no direct link between weight components and mean or variance components, so mixture density networks often learn seemingly arbitrary connections. We illustrate this experimentally in Section \ref{PR2results}, showing that an MDN trained to predict manipulator joint angles will use only a single mixture component for completely different joint angle predictions, somewhat unintuitively learning to change the mean and variance parameters instead of toggling between mixture components. This occurs because no structure forces mixture consistency in the network. 
 
\section{Switching density networks for hybrid control}

\begin{wrapfigure}{r}{0.6\textwidth}
    \vspace{0mm}
    \centering
    \includegraphics[width=0.6\textwidth]{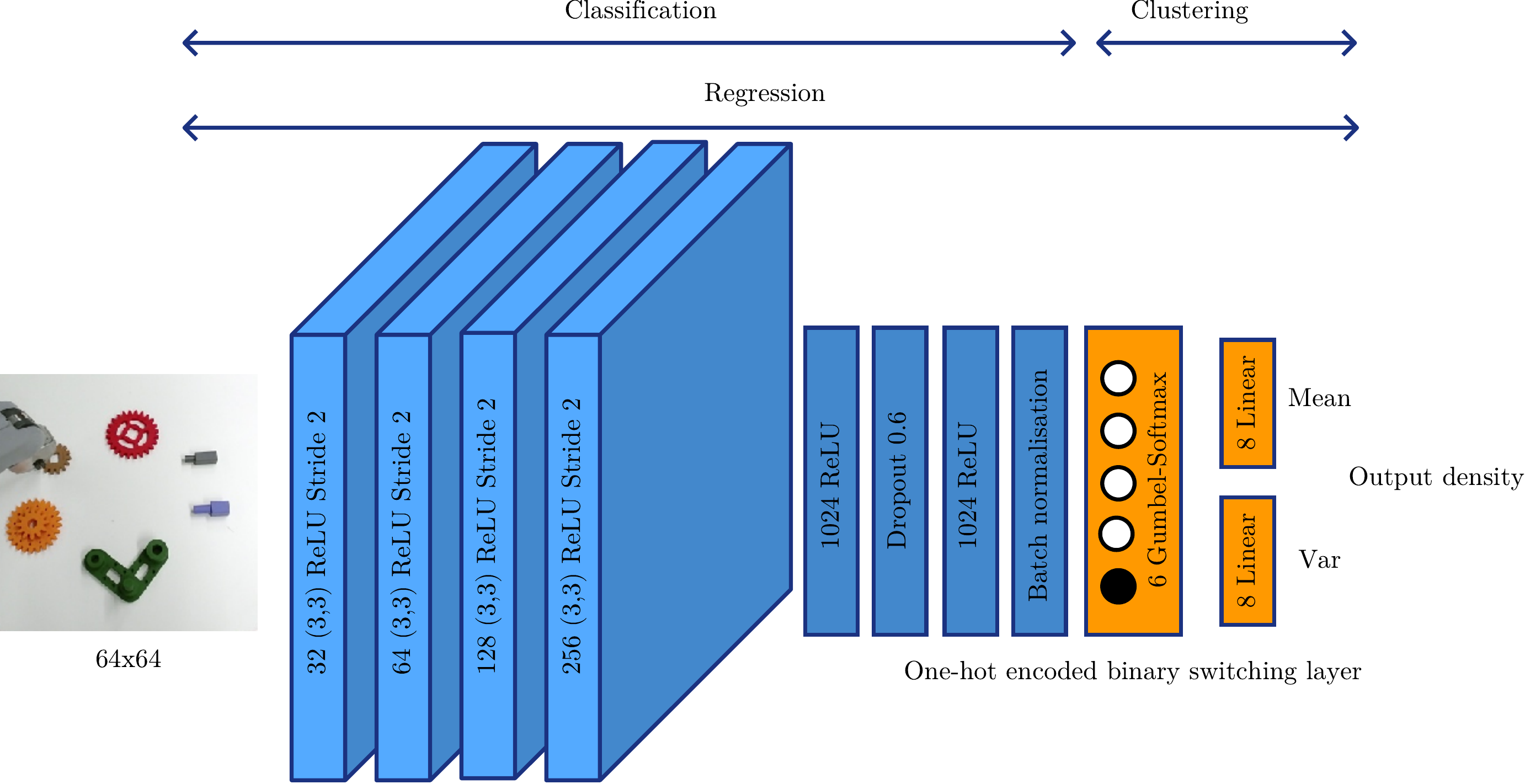}
    \caption{Switching density networks are discrete latent variable models that switch between output densities. This architecture is used for reaching controller identification in 8 DOF joint angle space.\label{fig:sdn}}
    \vspace{-5mm}
 \end{wrapfigure}

Switching state space models are typically learned using a multi-stage process. For example, Gaussian mixture models could be fit to robot state measurements, and perception networks trained to predict hidden states from image observations. Switching density networks attempt to learn hybrid systems like this jointly in an end-to-end fashion. More formally, given a hybrid system of $i=1\to N$ dynamical systems, each with parameters $\theta_i$,
\begin{align}
\dot{\mathbf{x}}_t = q(\mathbf{x}_t; i, \mathbf{\theta}_i),
\end{align}
we train a SDN to predict parameters $\mathbf{\theta}_i$, maximising the log likelihood of the distribution $\mathcal{N}(\dot{\mathbf{x}}_t|q(\mathbf{x}_t; i, \mathbf{\theta}_i),\Sigma_i)$, where $\Sigma_i$ denotes the measurement uncertainty. Figure \ref{fig:sdn} shows an example SDN architecture. A SDN is similar to a mixture density network, which typically consists of a neural architecture that predicts $K$ different sets of  distribution parameters, along with a set of discrete weights $K$. However, unlike mixture models, switching models only predict a single output distribution, which is conditioned on a $K$-dimensional one-hot encoded discrete latent variable. The final layer of a SDN is a fully connected layer with no bias parameters. When combined with the bottleneck switching layer, this produces a switching state as output.

Gradient-based learning with discrete latent variables is challenging, as backpropagation is unsuitable for non-differentiable layers. The Gumbel-softmax distribution \cite{jang2016categorical} approximates a categorical distribution using a temperature ($\tau$) controlled softmax function,
\begin{align}
y_i = \frac{\exp{((\log(\pi_i) + g_i)/\tau)}}{\sum_{j=1}^K \exp{((\log(\pi_i) + g_i)/\tau)}} \; \text{for} \; i = 1\hdots K,
\end{align}
with logits $\pi_i$ and i.i.d samples $g_i$ drawn from a Gumbel(0,1) distribution. As the temperature $\tau$ tends to 0, samples from the Gumbel-softmax distribution tend towards a one-hot encoded binary vector. As a result, neural model training using temperature annealing allows for backpropagation to be used to learn models with discrete latent parametrisations. 
 
 The Gumbel-softmax reparametrisation allows switching density networks to be trained with a discrete bottleneck layer, using stochastic gradient descent to minimise the negative log likelihood of a state given a network prediction conditioned on an input observation.
 
 In contrast to typical hybrid system identification problems, where system dynamics are identified, in a behaviour cloning setting we only observe the {\textit{closed-loop}} or controlled response of the system of interest, so need to infer switching controller models. As a particularly useful example, we choose to express the behaviour of a robot using a generative switching model comprising a sequence of PID controllers, motivated by the proportional control formulation of \citet{burke2019explanation}. PID control laws produce controller actions $\mathbf{u}_k$,
\begin{align}
\mathbf{u}_k = K_p (\mathbf{x}_k-\mathbf{\mu}) + K_i \sum_{l=1}^L(\mathbf{x}_{k-l}-\mathbf{\mu}) + K_d \frac{\mathbf{x}_{k}-\mathbf{x}_{k-1}}{\Delta_t},
\end{align}
using three error terms comprising a proportional gain $K_p$ acting on the error between state $\mathbf{x}_k$ at time step $k$ and a desired reference point $\mu$, an integral gain $K_i$ acting on the cumulative error between state $\mathbf{x}_k$ and a desired reference point $\mu$ over a window of time steps $l$, and a derivative gain $K_d$ acting on the change in state, over a time difference $\Delta_t$. 

In a behaviour cloning setting, we typically observe state-action pairs and train models to regress the appropriate action for a given state. However, if we assume that actions should be produced by proportional-integral-control laws, we can reformulate the regression problem as one of predicting the controller gains and reference points that generate observed actions. In complex systems, it may be the case that we require multiple PID control laws in different states and are required to switch between control laws. In this case, and assuming Gaussian observation noise, we can model the controller action,
\begin{align}
\mathbf{u}_k \sim K_p(\mathbf{z}_t, i_t)\left[\mathbf{x}_t-\mathbf{\mu}(\mathbf{z}_t, i_t)\right] + K_i(\mathbf{z}_t, i_t) \sum_{l=1}^L[\mathbf{x}_{k-l}-\mathbf{\mu}(\mathbf{z}_t, i_t)] +\nonumber\\ K_d(\mathbf{z}_t, i_t) \frac{\mathbf{x}_{k}-\mathbf{x}_{k-1}}{\Delta_t} +  \mathcal{N}(\mathbf{0},\mathbf{\Sigma}(\mathbf{z}_t, i_t)), \label{prop_switch}
\end{align}
using a hybrid system of $i$ distinct control law parameters and reference points conditioned on some sensor data. Here, each controller is parametrised by a set of gains, $K_p(\mathbf{z}_t, i_t),K_i(\mathbf{z}_t, i_t),K_d(\mathbf{z}_t, i_t)$, and goal configuration states, $\mathbf{\mu}(\mathbf{z}_t, i_t)$. Controller action is measured subject to uncertainty, $\mathbf{\Sigma}(\mathbf{z}_t, i_t)$. Sequencing a number of these controllers allows a robot to transition through the set of states required to solve many manipulation and navigation tasks.

Given this model and a demonstration sequence of state and observation pairs, our goal is to learn to identify $K$ suitable sub-controllers that make up the demonstrated robot behaviour. For the Gaussian proportional controller model of (\ref{prop_switch}), we train switching density networks using the log-likelihood ($\mathcal{LL}$) objective:
\begin{align}
\theta^* = \text{argmin}_\theta -\mathcal{LL}(\mathbf{u}_k|K_p(\mathbf{z}_t;\theta),K_d(\mathbf{z}_t;\theta),K_i(\mathbf{z}_t;\theta),\mathbf{\mu}(\mathbf{z}_t;\theta),\mathbf{\Sigma}(\mathbf{z}_t;\theta)),
\end{align}
In many cases, there may be multiple possible gains and reference points that produce an observed controller action. In practice, we address this by only predicting a subset of the controller parameters, through sensible weight initialisation in the final layer of the SDN and by relying on persistent excitation in the demonstration sequence.

Unfortunately, training a switching density network frequently results in mode collapse. Here, all probability mass becomes concentrated in a single class, and the network merely learns to regress the mean state. We remedy this by using an additional cross-entropy loss term, which serves as a batch-level regulariser. Here, we assume that all categories are likely to occur at a similar rate in a given batch, and minimise the cross-entropy between the average Gumbel-softmax distribution over a batch $\hat{y}_i$ and a uniform distribution, 
\begin{align}
\theta^* = \text{argmin}_\theta -\mathcal{LL}(\mathbf{u}_k|K_p(\mathbf{z}_t;\theta),K_d(\mathbf{z}_t;\theta),K_i(\mathbf{z}_t;\theta),\mathbf{\mu}(\mathbf{z}_t;\theta),\mathbf{\Sigma}(\mathbf{z}_t;\theta)) - \sum_{i=1}^K \frac{1}{K}\text{log}(\hat{y}_i).
\end{align}
This ensures that on average no individual category in the switching layer becomes dominant, and helps to counter mode collapse.

\section{Experimental Results}

We evaluate SDN controller identification in two distinct domains. The first, an inverted pendulum, is a canonical hybrid continuous control problem, while the second, a set of visuomotor manipulation tasks, illustrates the applicability of SDNs to higher dimensional input and output spaces. For both experiments, we specify the exact number of controllers to be identified. In practice, should this number not be known, more controller switches could be specified, to allow for redundancy.

\subsection{Balancing an inverted pendulum}

We demonstrate the use of SDN PID controller laws on a simulated, under-actuated inverted pendulum. Closed-loop control of an inverted pendulum can be accomplished using a hybrid system with three key modes \cite{Kuipers02}, that derive from fundamental properties of the physics of this system. In the pump mode, energy is injected to the system, such that it can be swung up. In the spin mode, energy is removed from the system through damping action. Finally, when the pendulum is near vertical, a balancing controller can be used to maintain the pendulum in an upright position. Control in each mode can be provided by a proportional control law, with the energy in the system used to determine which control law to apply. 

\begin{figure}
    \centering
    \begin{subfigure}{0.333\textwidth}\includegraphics[width=\textwidth]{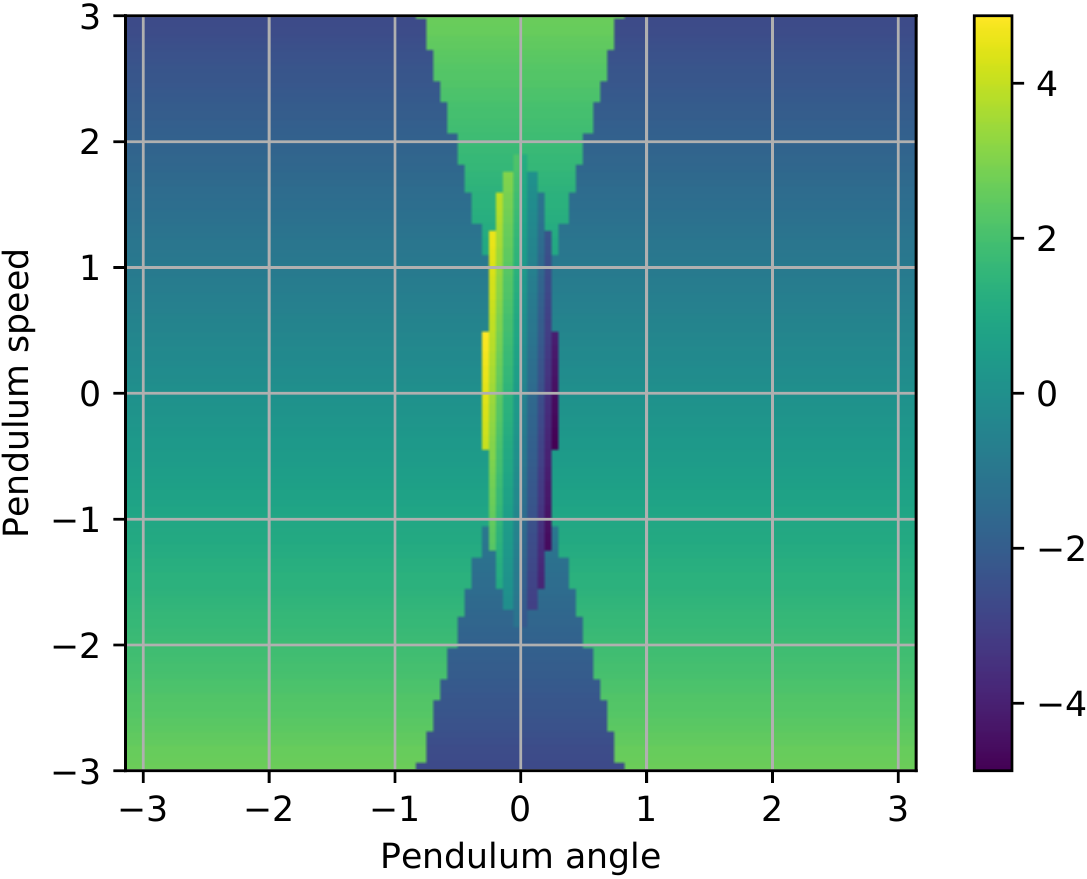}\caption{Hybrid controller}\end{subfigure} \begin{subfigure}{0.276\textwidth}\includegraphics[width=\textwidth]{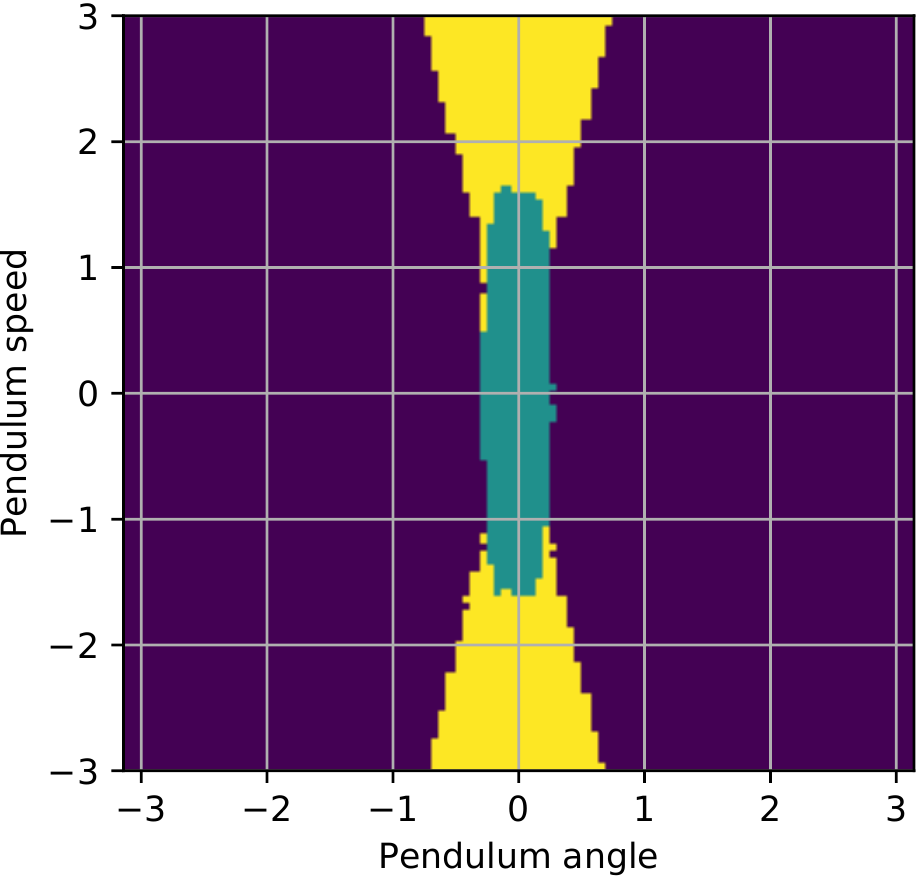}\caption{Sub-controller regions}\end{subfigure} \begin{subfigure}{0.333\textwidth}\includegraphics[width=\textwidth]{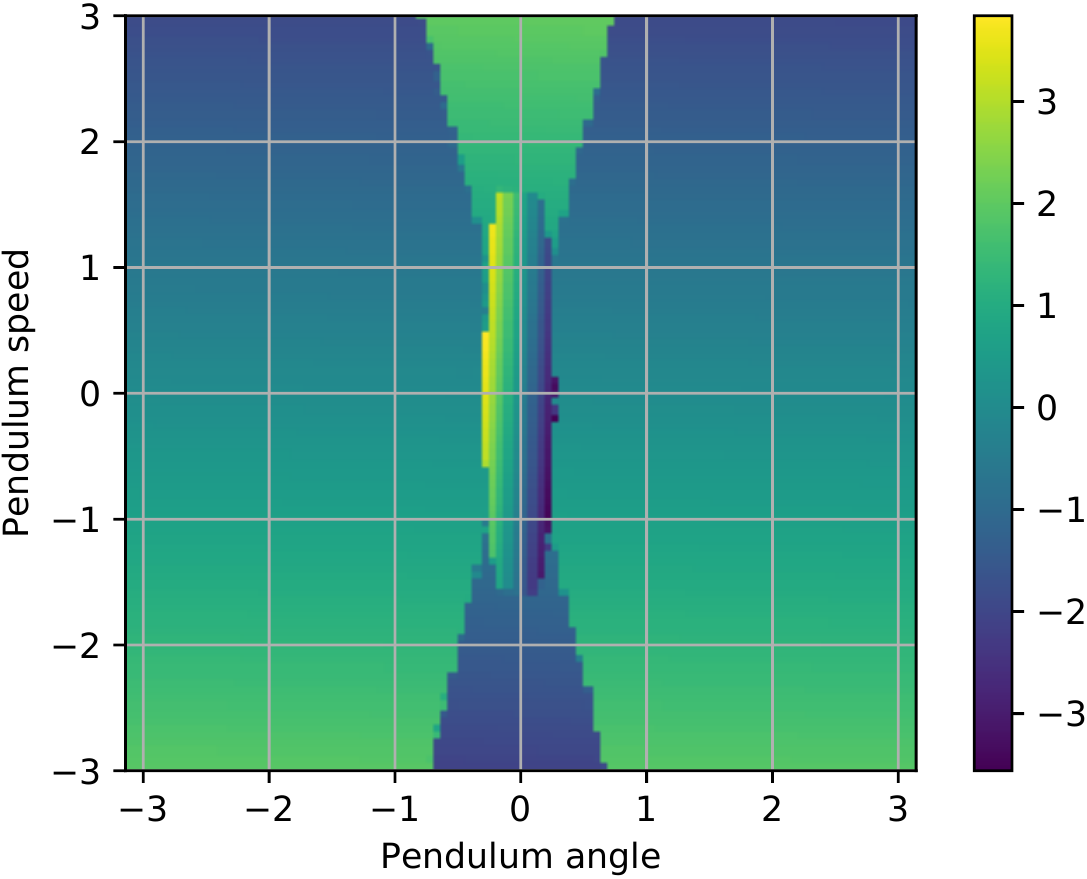}\caption{SDN response}\end{subfigure}
    \caption{Controller responses for inverted pendulum learned using the SDN closely match those of the ground truth hybrid controller used for demonstration. Importantly, the SDN correctly identifies the regions in state space in which each controller should be applied.\label{fig:pendulum_response}}
\end{figure}

We train a SDN (3 fully connected layers of 16 neurons, and 3 switching states) to predict the proportional controller gains (no reference points are needed as the pendulum goal state is known) for the three controllers described above, using 10,000 state-action pairs provided by the original hybrid controller \cite{Kuipers02}. Figure \ref{fig:pendulum_response} shows the controller response in different regions of the state space, with the three controller regions clearly visible. Importantly, the figure shows that SDNs learn to identify the regions in which each of these controllers should be applied, in addition to the required sub-controller parameters. Direct behaviour cloning using a neural network is still effective, but does not provide the same level of interpretability as the learned hybrid control system.

\begin{wraptable}{l}{0.4\textwidth}
    %\centering 
    \vspace{-4mm}
    \caption{Pendulum experiments\label{tab:pendulum_sdn}\vspace{-2mm}}
    \begin{tabular}{|l|l|l|l|l|l|}
    \hline
    & Average reward \\
    \hline
         Hybrid controller &  $-0.812 \pm 0.514$  \\
    \hline
         SDN PID &  $-0.857 \pm 0.621$ \\
    \hline
         Fully connected & $-0.850 \pm 0.538$  \\
    \hline
    \end{tabular}
    \vspace{-3mm}
\end{wraptable}The latent structure can be used to interpret and reason about the underlying dynamics and physics of the controlled system and environment. For example, by analysing the regions in which sub-controllers operate (Figure  \ref{fig:pendulum_response}), along with the inferred controller parameters, we can see that the pendulum requires positive feedback control in the blue region, indicating the presence of a system with a stable equilibrium point, and negative feedback control in the blue-green region, indicating an unstable equilibrium point.

Table \ref{tab:pendulum_sdn} shows the average reward ($\theta^2 + 0.1\dot{\theta}^2 + 0.001u^2$, where $(\theta, \dot{\theta})$ denotes pendulum angle and velocity, and $u$ the control) obtained over 1,000 randomly initialised tests using the ground truth hybrid controller, and controllers learned using a SDN. Behaviour cloning using a baseline, fully connected network with no structure performs similarly to a SDN, but the latter is more interpretable, as specific control laws can be linked to regions in state space, allowing for more detailed controller analysis. Importantly, the inferred control law parameters are close to the ground truth values, indicating that the SDN has successfully identified the underlying control laws demonstrated. 

\subsection{Identifying PR2 manipulation controllers}
\begin{figure}
    \centering \vspace{-1mm}
    \includegraphics[width=0.88\textwidth]{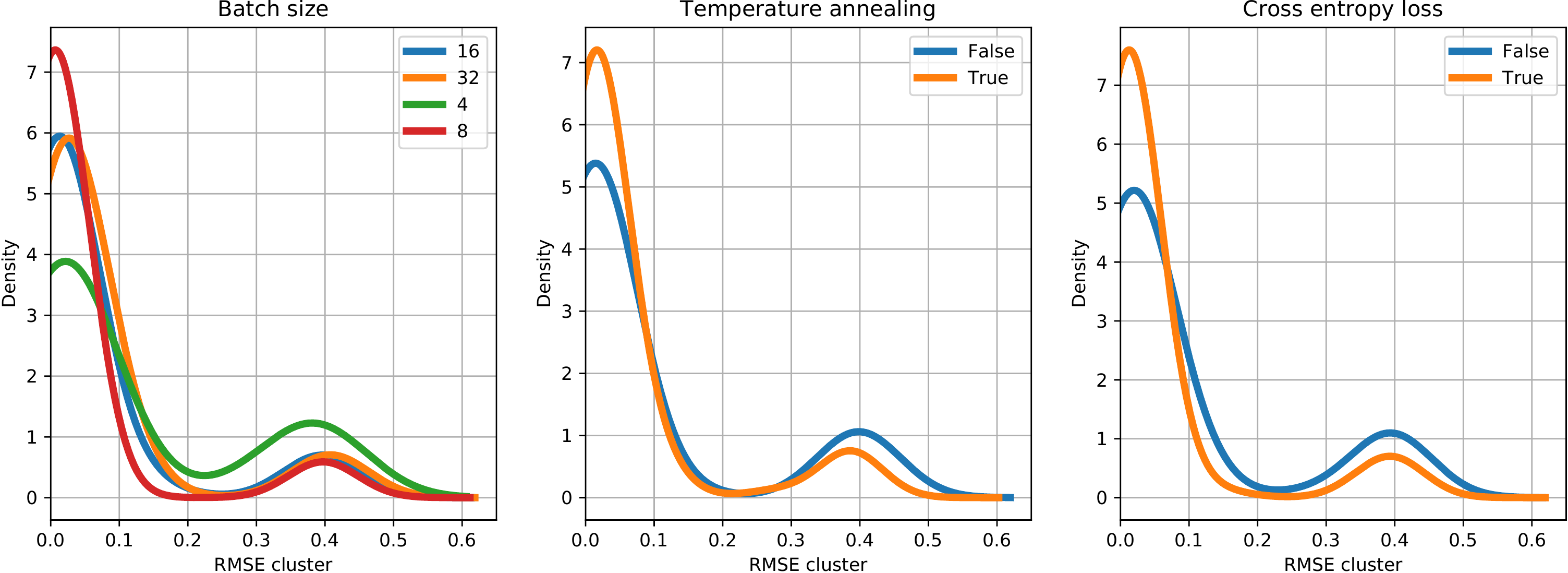}
    \caption{Kernel density estimates of root mean square error between true controller goals and PR2 goal predictions highlight the value of cross entropy regularisation and temperature annealing.\label{fig:rmse_kde_cluster}}
    \vspace{-5mm}
\end{figure}
\label{PR2results}

We also evaluate the use of switching density networks (Architecture in Figure \ref{fig:sdn}) for controller goal identification using an inspection task with a PR2 robot. Here, the PR2 is required to repeatedly reach to a series of components. We hard coded this behaviour and collected approximately 2,000 images and corresponding joint angle (8 dimensions) and velocity measurements, while the PR2 repeated this process 10 times. Our goal is to learn to identify the sub-controllers that make up this sequence, and train a model to predict these controller parameters from image observations. We split this set into two, with the first 1,000 frames used for training and the remainder for testing.

\begin{wraptable}{l}{0.18\textwidth}
    \centering
    \vspace{-5mm}
    \caption{RMSE\label{tab:mdn_sdn}\vspace{-2mm}}
    \begin{tabular}{|l|l|}
    \hline
         SDN &  $0.8^\circ$\\
         %Batch size 8, annealing and cross entropy
    \hline
        CNN &  $1.47^\circ$\\
    \hline
         MDN &  $3.95^\circ$\\
         % Batch size 32, annealing and no cross entropy
    \hline
    \end{tabular}
    \vspace{-2mm}
\end{wraptable}Here, we assume that the controller gains are known, and only learn to predict controller reference points. Figure \ref{fig:rmse_kde_cluster} shows the distribution over the root mean square error in predicted joint angle goals for each frame in the test set. Results are provided for varying batch sizes, using varying Gumbel-softmax temperature annealing schedules, and with or without the proposed cross-entropy loss. Experiments were repeated 10 times for each parameter setting combination.

\begin{figure}
    \centering
    \begin{subfigure}{0.5\textwidth}\includegraphics[trim=4cm 1cm 2cm 0 ,clip,width=\textwidth]{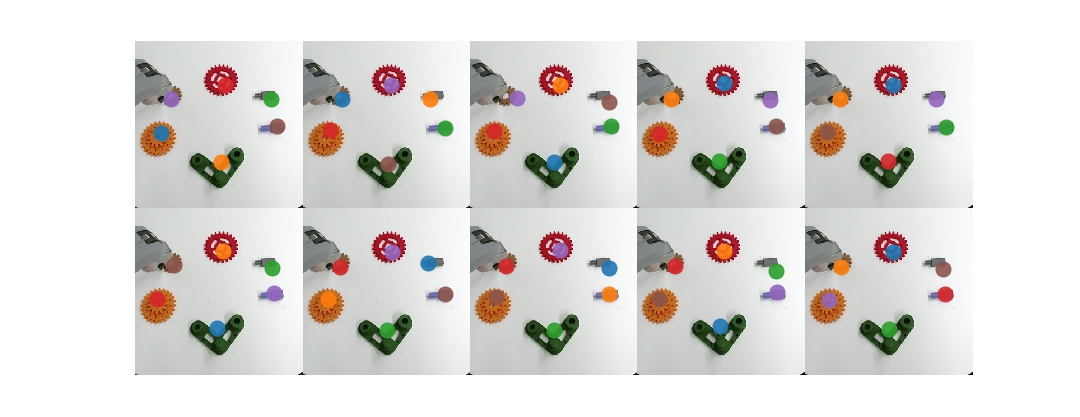}\caption{Cross entropy,  temperature}\end{subfigure}\begin{subfigure}{0.5\textwidth}\includegraphics[trim=4cm 1cm 2cm 0 ,clip,width=\textwidth]{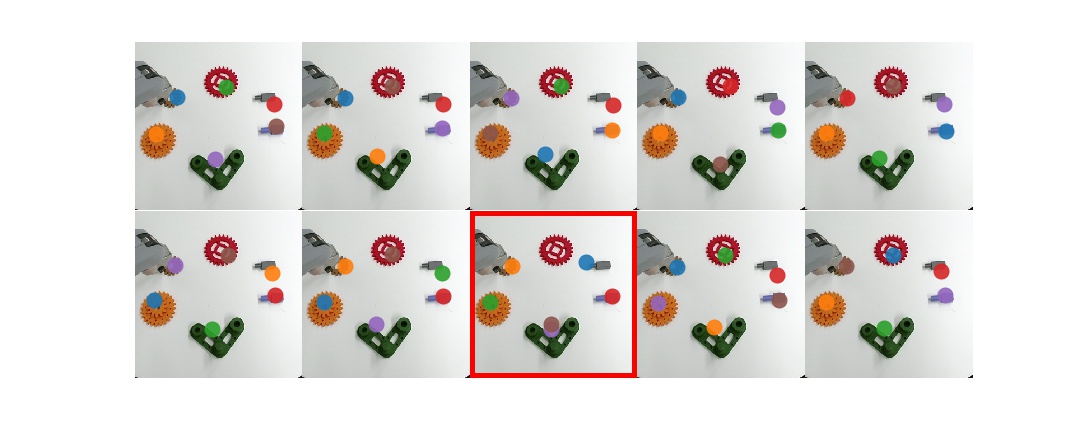}\caption{Cross entropy, no temperature}\end{subfigure}
    \begin{subfigure}{0.5\textwidth}\includegraphics[trim=4cm 1cm 2cm 0 ,clip,width=\textwidth]{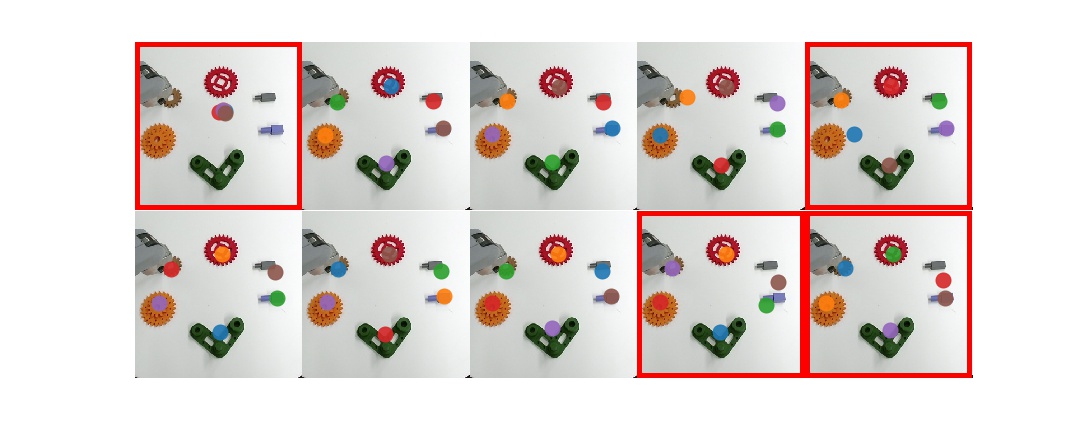}\caption{No cross entropy, temperature}\end{subfigure}\begin{subfigure}{0.5\textwidth}\includegraphics[trim=4cm 1cm 2cm 0 ,clip,width=\textwidth]{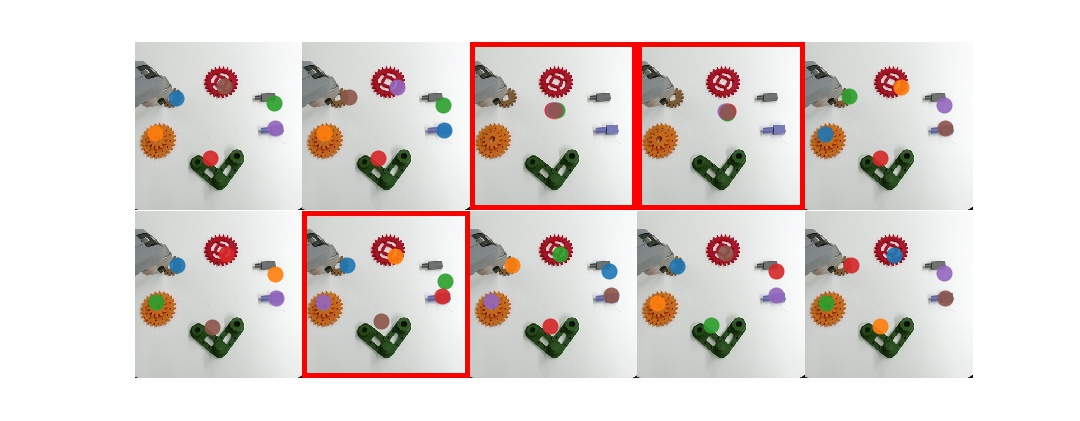}\caption{No cross entropy, no temperature}\end{subfigure}
    \caption{Projected joint angle goals (dots) identified using the SDN highlight the importance of cross entropy regularisation. Failed goal identification is indicated using a red border.\label{fig:sdn_goals}}
    \vspace{-5mm}
\end{figure}

It is clear that both temperature annealing and the cross-entropy loss are required for stable training. Batch size is a proxy for the rate of temperature annealing, since temperature was annealed at each epoch step, but also affects the cross-entropy loss term. If the batch size is too low, the assumptions governing the cross entropy regularisation loss are less likely to be true. This effect is clearly visible for the batch size of 4. Figure \ref{fig:sdn_goals} shows the projection of the detected joint angle goals into the image plane for experimental runs using the various parameter configurations. When trained with both a cross-entropy loss and temperature annealing the SDN successfully identifies the inspection goals comprising this task. Without the cross-entropy loss, the network is vulnerable to mode collapse, where predicted goals regress to the mean. 

\begin{wrapfigure}{r}{0.65\textwidth}
    \centering \vspace{-5mm}
    \includegraphics[width=0.65\textwidth]{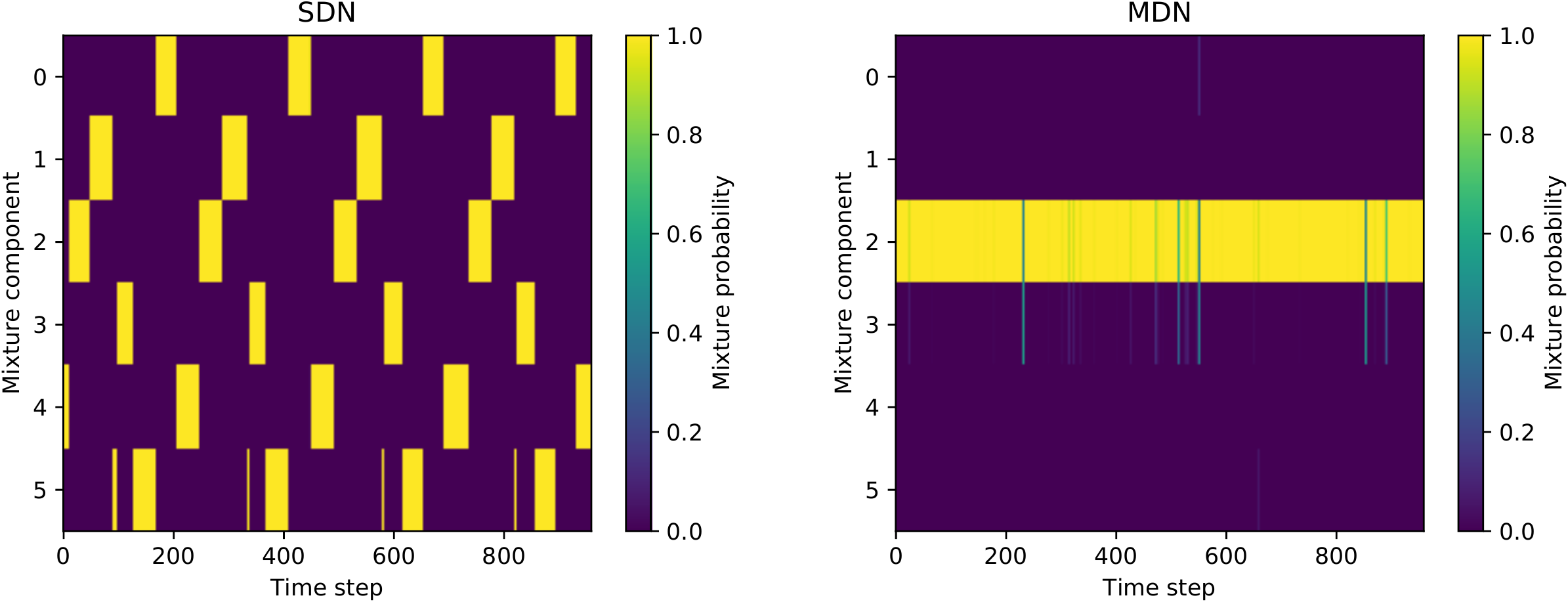}
    \caption{Unlike the SDN, MDNs do not necessarily use different mixture components for similar regressions.\label{fig:mixture_seq}}
    \vspace{-3mm}
\end{wrapfigure}
It is important to note the difference between switching and mixture density networks. The former enforces a hierarchical latent model structure, while the latter places no constraints on the mapping between mixture components and distribution outputs. This can be observed when the distribution of the mixture weights is shown for the test sequence for switching and mixture density networks, when both are trained to predict controller goals. SDNs implicitly learn the latent task behaviour, while MDNs simply regress using arbitrary paths through the network. Table \ref{tab:mdn_sdn} shows the root mean square errors over all joints for the best performing parameter settings on the inspection task. The MDN performs substantially worse than the SDN, which captures the inherent switching structure of the inspection task. In general, larger batch sizes improve MDN results, as does temperature annealing, but cross entropy regularisation has little effect. Figure \ref{fig:mixture_seq} shows the trace of mixture component densities, which highlights the fact that SDNs learn intrinsically meaningful grounded predictions, while MDNs rely on an arbitrary mapping between input and output. 
\begin{figure}
    \centering
    \includegraphics[width=\textwidth]{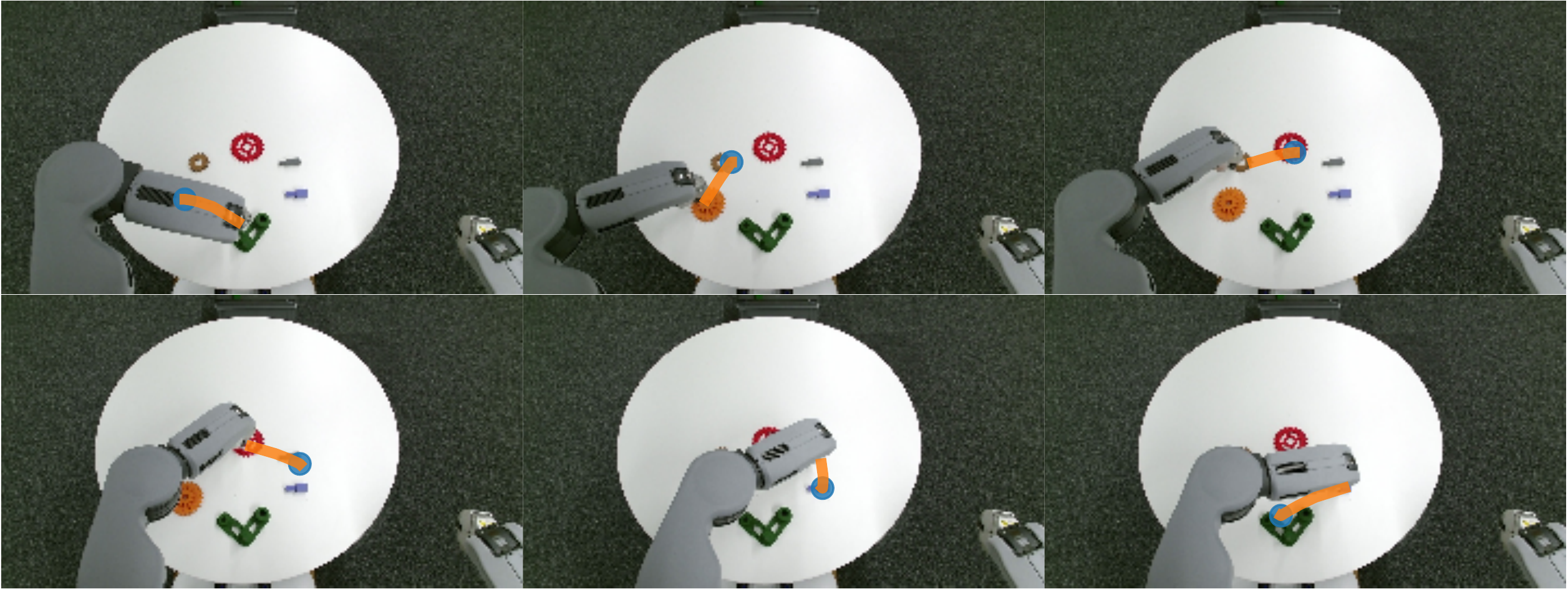}
    \caption{Inferred controller rollouts (joint angles projected into the image plane) are obtained by predicting the controller goal state for the given image using the SDN, and then using the associated PID controller to generate a trajectory.\label{fig:trace}}
    \vspace{-5.5mm}
\end{figure}

Figure \ref{fig:trace} shows the projected controller rollouts obtained using the SDN. It is clear that the SDN has identified the appropriate switching points and control strategies need to replicate the inspection task. This is particularly useful for learning from demonstration, as this options discovery allows for the inclusion of higher level reasoning about the inspection plan being followed, and the identification of implicit constraints or search patterns followed by the demonstrator.

\begin{wrapfigure}{r}{0.55\textwidth}
    \vspace{-1mm}
    \centering 
    \includegraphics[width=0.55\textwidth]{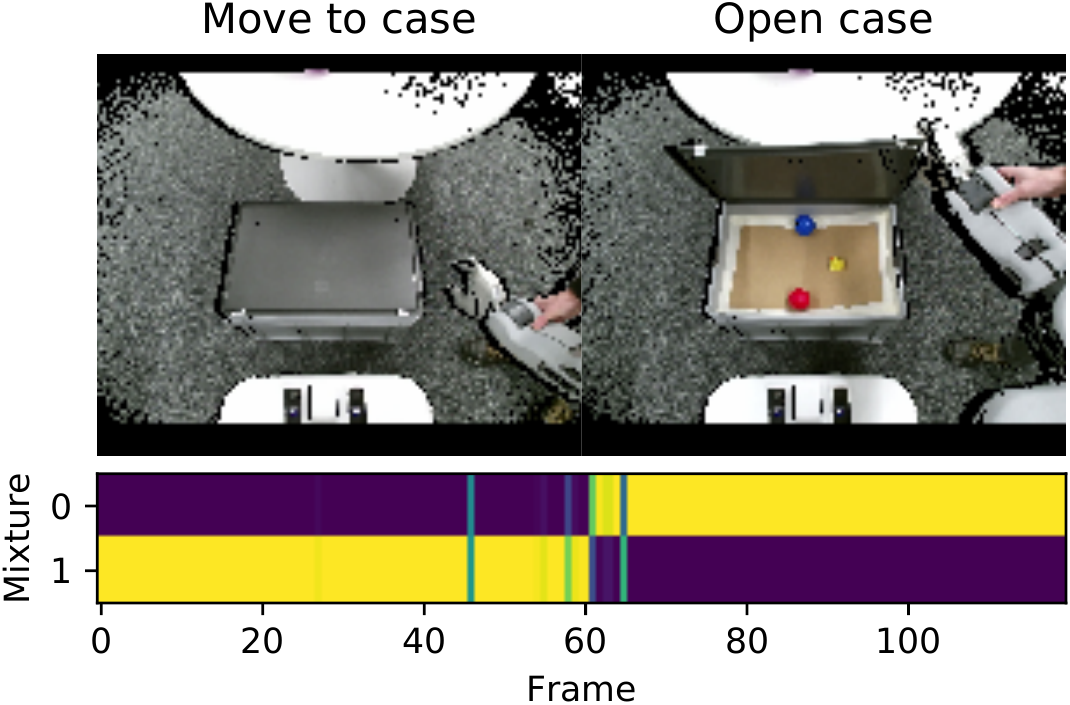}
    \caption{Two controllers are identified for a kinesthetically demonstrated suitcase opening task. \label{fig:suitcase}}
    \vspace{-4mm}
\end{wrapfigure}

Although seemingly simple, even contact rich behaviours can be expressed using PID control laws. For example, when we applied a SDN to a kinesthetically demonstrated suitcase opening task, we discovered two primary controllers (Figure \ref{fig:suitcase}), one moving beneath the case lid, and a second that opened it by moving to a goal state above the case. Importantly, the SDN allows for the use of vision to determine which controller to apply.

Empirically, we have found that fully connected networks were easier to train (faster convergence) and perform slightly better than SDNs on tasks that can be solved using a smooth non-linear controller (pendulum), but that SDN's are easier to train and provide substantial improvements over direct CNNs on tasks with a clear switching structure (inspection task). Our hypothesis is thus that SDNs are best suited to hybrid systems, which cover a broad set of processes. However, we believe that the increased training difficulty in the first case is made up for by the substantial interpretability gains that can be obtained by discretising a process using an SDN.

\section{Conclusion}
\label{sec:conclusion}

This work has introduced an approach for end-to-end hybrid system identification using switching density networks and generalised state-space control laws, in the context of behaviour cloning. Switching density networks are harder to train than their fully connected counterparts, but this work has shown empirically that the addition of a cross-entropy regularisation term stabilises training. Hybrid systems are frequently used in robotics, and PID controllers are a trusted and well understood control paradigm, used widely across domains and disciplines. Although demonstrated using PID control laws, the proposed approach allows for hybrid system identification using other controller families. Importantly, jointly inferring sub-controllers and the states in which they are applied allows for reasoning about implicit constraints and the physical properties of systems and environments.

This work has shown that SDNs can be used to perform goal identification in a visuomotor manipulation tasks and inverted pendulum controller discovery, identifying compositional control strategies that can be directly used for explainable control. We have contrasted these with MDNs and fully connected neural networks, which are unable to learn interpretable latent representations. Moreover, hybrid system identification using SDNs allows for the re-use of learned policies in downstream tasks, through hierarchical reinforcement learning or options scheduling. Future work will involve exploring options scheduling with sub-controllers learned using SDNs.

%It should be noted that switching density networks are challenging to train, and vulnerable to mode collapse. The proposed regularisation strategy remedies this to an extent, but we found it difficult to learn switching state space models where the controllers came from completely different families, particularly when the number of switching components is limited. Further work on improved training and network intitialisation strategies that allow for the use of arbitrary control strategies would be beneficial.

%===============================================================================

% The maximum paper length is 8 pages excluding references and acknowledgements, and 10 pages including references and acknowledgements

%\clearpage
% The acknowledgments are automatically included only in the final version of the paper.
\acknowledgments{This work is supported by funding from the Turing Institute, as part of the Safe AI for surgical assistance project. We
are particularly grateful to the Edinburgh RAD group for valuable discussions and recommendations.}

%===============================================================================

\bibliography{example}  % .bib

\end{document}